\RequirePackage{xcolor}
\documentclass{article}

\usepackage{arxiv}

\usepackage{amsmath}
\usepackage{amssymb}
\usepackage{adjustbox}
\usepackage{tabularx,ragged2e}
\usepackage{epsfig}
\usepackage{graphicx}
\usepackage{multirow}
\usepackage{spreadtab}
\usepackage{booktabs}
\usepackage{xspace}
\usepackage[utf8]{inputenc}
\usepackage[ruled]{algorithm2e}
\usepackage{algorithmic}

\fancypagestyle{alim}{\fancyhf{}\fancyfoot[L]{\small This is the accepted version of a workshop paper presented at the British Machine Vision Conference (BMVC, Cardiff, United Kingdom, September 9-12, 2019)}}

\newcommand{\partitle}[1]{\bigbreak\noindent\textbf{#1}}
\newcommand{\partitlenobreak}[1]{\noindent\textbf{#1}}
\newcommand{\normeuc}[1]{\left\lVert {#1} \right\rVert_2}
\newcommand{\bfnot}[1]{\mathbf{#1}}
\newcommand{\model}{~SafeCritic\xspace}
\newcommand{\reward}{Critic~}
\title{SafeCritic: Collision-Aware Trajectory Prediction}

\author{
Tessa ~van der Heiden\thanks{BMW Group, 
Landshuter Str. 26, 85716 Unterschlei{\ss}heim, Germany} \\
\texttt{tessa.heiden@bmw.de}
\And Naveen Shankar~Nagaraja\textsuperscript{*} \\
\texttt{naveen-shankar.nagaraja@bmw.de}\\
\And Christian Wei{\ss}\textsuperscript{*} \\
\texttt{christian.wc.weiss@bmw.de}
\And
Efstratios Gavves\thanks{ University of Amsterdam,
Science Park 904, 1012 WX Amsterdam, Netherlands} \\
\texttt{egavves@uva.nl}
}

\def\eg{\emph{e.g}}

\def\etal{\emph{et al}}
\def\ie{\emph{i.e}}

\DeclareMathOperator*{\argmin}{arg\,min}

\begin{document}
\maketitle
\thispagestyle{alim}
\begin{abstract}
Navigating complex urban environments safely is a key to realize fully autonomous systems. Predicting future locations of vulnerable road users, such as pedestrians and cyclists, thus, has received a lot of attention in the recent years. While previous works have addressed modeling interactions with the static (obstacles) and dynamic (humans) environment agents, we address an important gap in trajectory prediction.
We propose \model, a model that synergizes generative adversarial networks for generating multiple ``real'' trajectories with reinforcement learning to generate ``safe'' trajectories. The Discriminator evaluates the generated candidates on whether they are consistent with the observed inputs. The Critic network is environmentally aware to prune trajectories that are in collision or are in violation with the environment. The auto-encoding loss stabilizes training and prevents mode-collapse. We demonstrate results on two large scale data sets with a considerable improvement over state-of-the-art. We also show that the Critic is able to classify the safety of trajectories. 

\end{abstract}


\section{Introduction}\label{sec:intro}
% What is the task?
Forecasting future locations of humans is an essential task to develop safe autonomous agents that interact with people.
Autonomous systems such as  household and factory robots, security systems and self driving cars need to know where people will be given their previous movement to plan safe paths. 

% What are the challenges?
Predicting people's future movement is challenging as the target location of a person cannot be known.
The static and dynamic layout of the environment, be it a road network or an indoors map complicates the forecasting.
Static obstacles place restrictions beyond their narrow definition of spatial extent (pedestrians will tend not to walk between closely situated parked cars).
Considering dynamic obstacles like other pedestrians, bicycles or moving cars, one must not only minimize direct future collisions but also take into account hidden social rules (keeping to the right, respecting personal space).
This becomes more complex since there are multiple unknown target locations each of the dynamic obstacles might be heading to.
Consequently, predicting all the possible and plausible  future trajectories is a multi-modal problem. 
Respecting the hidden rules and addressing the challenges is key for systems that are not just accurate but also safe.

% How do current researchers solve this problem?
Computationally tractable generative models has led to a great number of methods~\cite{gupta2018social, sadeghian2018sophie, lee2017desire} that model the multi-modality of future trajectory prediction.
While these models focus on producing accurate future trajectories, they do not take into account the safety of the trajectories explicitly in terms of collisions.
As a result, the generated paths might be \textbf{unsafe}, wherein they collide with other road users, or even \textbf{infeasible}, i.e., they go through static scene elements like buildings, cars or trees.
Consequently, these trajectories cannot easily be trusted in practice.

%Modelling future trajectories of road users by minimizing displacement errors from ground truth alone is not a sufficient objective. Importantly, the predicted trajectories must also be realistic, namely to not produce \textbf{infeasible and unsafe} paths where road users collide with each other or with static scene elements like trees and cars. 

In this work we argue that one should take into account safety, when designing a model to predict future trajectories.
The problem is that future trajectories are generated online and cannot be predetermined.
Thus, applying standard and differentiable supervised learning is not straightforward.
Even if one could minimize collisions with static elements and the ground truth trajectories, the model must learn to minimize collisions with the generated, and therefore unseen, future trajectories.
Thus, in the absence of a clear learning objective, we rely on Generative Adversarial Networks (GAN)~\cite{goodfellow2014generative} for the generative modelling and cast the learning within the Inverse Reinforcement Learning framework~\cite{IRL}.
Inverse Reinforcement Learning guides the model towards optimizing an unknown reward function and shares many similarities with GANs.
Here, our unknown reward function accounts for learning safe and ``collision-free'' future trajectories that have not even been generated yet.

This work has three contributions: (a) to the best of our knowledge, we are the first to propose generating ``safe'' future trajectories, minimizing collisions with static and dynamic elements in the scene,
(b) the generated trajectories are not only safer but also more accurate. 
The reason is that the model learns to prune out unlikely trajectories and generalize better.
In fact, we show that the remaining few collisions in the trajectories strongly correlate with ground truth, thus showing that the model learns to also anticipate collisions. %estimates the likelihood of future collision. 
and (c), we rely on attention mechanism for spatial relations in the scene to propose a compact representation for modelling interaction among all agents.
The representation, which we refer to as attentive scene representation, is invariant to the number of people in the scene and thus is scalable.

%-------------------------------------------------------------------------
\section{Related work}\label{sec:rel_work}

\partitlenobreak{Future trajectory prediction.}
Predicting future trajectories of road users has recently gained an increased attention in the literature.
In their pioneering work Alahi \etal \cite{dasgupta_cvpr16} define future trajectory prediction as a time series modelling problem and proposed SocialLSTM, which aggregates information of multiple agents in a single recurrent model. 
While SocialLSTM is a deterministic model, Gupta \etal \cite{gupta2018social} proposed SocialGAN to learn to generate multi-modal future trajectories.
To make sure that novel future trajectories are sampled, the encoder and decoder were trained using a Conditional Generative Adversarial Network \cite{gauthier2014conditional,mirza2014conditional} learning framework, leading to impressive results. 

% Taking obstacles into account
Extending on SocialGAN, SoPhie \cite{sadeghian2018sophie} incorporate semantic visual features extracted by deep networks, combined with an attention mechanism.
Inspired by~\cite{sadeghian2018sophie}, DESIRE \cite{lee2017desire} also relies on semantic scene context and introduces a ranking and refinement module to order the randomly sampled trajectories by their likelihood.
Further, \cite{rhinehart2018r2p2} develop R2P2, which is a non-deterministic model trained by modified cross-entropy objective to balance betwee the distribution and demonstrations. SocialAttention \cite{vemula2018social} connects multiple RNNs in a graph structure to efficiently deal with the varying amount of people in a scene. 

% Alternative to deep learning
Future trajectory prediction models beyond the deep learning have also been explored.
The Social Force \cite{helbing1995social} is one of the first models of pedestrian trajectory prediction that was successfully applied in the real-world \cite{sud2008real}. Interacting Gaussian Process (GP) based methods \cite{trautman2010unfreezing} model the trajectory of each person as a GP with interaction potential terms that couples each person's GP.
An extensive array of comparisons is available in~\cite{becker2018evaluation}, based on the TrajNet~\cite{sadeghiankosaraju2018trajnet}  benchmark.

Here, we adopt a modified Conditional Generative Adversarial Network framework~\cite{gauthier2014conditional,mirza2014conditional} to generate novel future trajectories from observed past ones.
Compared to~\cite{gupta2018social, sadeghian2018sophie, sadeghian2018sophie, lee2017desire} our focus is to generate trajectories that are not just accurate but also lead to minimum collisions and thus are safe.
Safe trajectories are different from trajectories that try to imitate the ground truth~\cite{rhinehart2018r2p2}, as the latter may lead to implausible paths, \eg, pedestrians going through walls.

\partitle{Generative Adversarial Networks + Reinforcement Learning.}
To model safe trajectories we propose a model that combines Generative Adversarial Networks
Generative Adversarial Networks~\cite{goodfellow2014generative} and Inverse Reinforcement Learning~\cite{IRL}. \cite{goodfellow2014generative} are generative models that rely on the min-max optimization of two competing sub-models, the generator and the discriminator.
As standard GANs do not have an encoder, Conditional GANs \cite{gauthier2014conditional, mirza2014conditional} include an encoder, which can be used to generate novel samples given observations.
Finn \etal \cite{finn2016gan_irl} show that GANs resemble Inverse Reinforcement Learning~\cite{IRL} in that the discriminator learns the cost function and the generator represents the policy. 
Inspired by~\cite{de2018molgan, guimaraes2017objective} who showed the potentials of using external simulators as critic objectives~\cite{konda2000actor}, we propose to use a critic network to model collision minimization by Inverse Reinforcement Learning.
To prevent mode collapse, an additional auto-encoding loss is considered.

% % Actor critic methods
% Recently, De Cao and Kipf \cite{de2018molgan} and Guimaraes \etal \cite{guimaraes2017objective} proposed MolGAN and ORGAN models for drug discovery respectively, both based on the SeqGAN \cite{yu2017seqgan}.
% They incorporate an external simulator to their GAN formulation, which indicates the validity of the sampled molecule.
% To accommodate the reward signal from the external simulator, MolGAN and ORGAN propose an additional critic objective to the discriminator network of the GAN. 

% In this work we propose a reinforcement learning component, the \emph{\reward Module}, for our future trajectory generator.
% Specifically, the \reward module is combined with the conditional GAN and is trained to minimize future collisions. We employ the DDPG model~\cite{sutton1998introduction} which incorporates a deterministic actor and critic network, instead of applying the REINFORCE algorithm \cite{williams1992simple} which returns unbiased but high variance gradients \cite{konda2000actor}. 
% While the GAN discriminator ensures the generated trajectories seem realistic, the \reward ensures the generated trajectories are safe and do not collide with dynamic or static elements in the scene. In addition to the reward signal and adversarial loss, the auto-encoding loss is used to stabilize training and preventing mode-collapse. This is similar to GAIL \cite{ho2016generative}, that uses a GAN to map expert trajectories to a policy. In this work, instead of computing the error signal between ground truth and all generated samples, it is computed from the best sample.

\section{\model}\label{sec:method}

\begin{figure*}[t]
\begin{center}
%\fbox{\rule{0pt}{2in} \rule{.9\linewidth}{0pt}}
   \includegraphics[trim=10 80 30 10,clip,width=\textwidth]{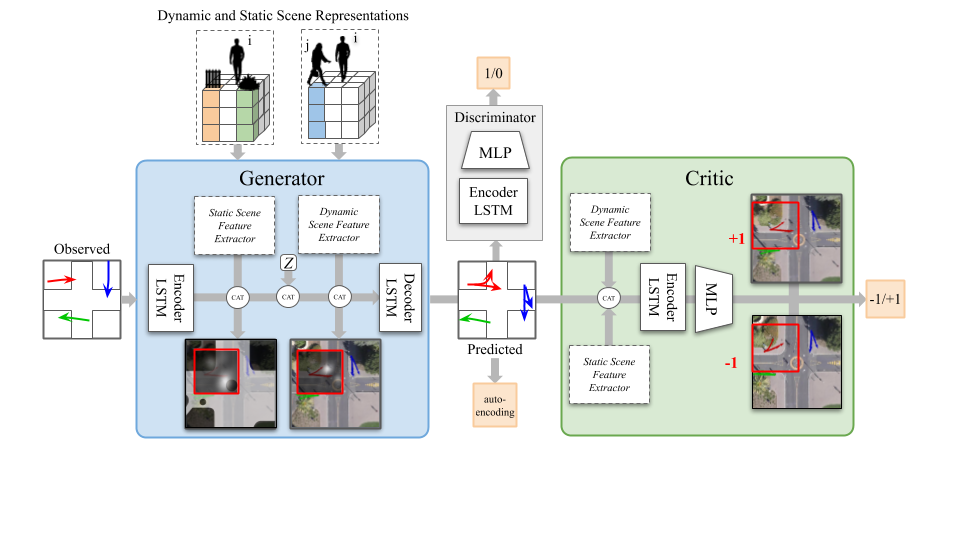}
\end{center}

  \caption{\small{The \model network consisting of a \textbf{Generator} trained by \textbf{Discriminator} and \textbf{\reward} networks. The Generator observes each trajectory together with the dynamic and static scene and computes a future trajectory. The \reward and discriminator score the predictions. The orange boxes are the three different losses for the Generator. The \textbf{Dynamic Scene} Representation are the hidden states of each neighbor $j$ and embedded into a 3D tensor for each target agent $i$. The \textbf{Static Scene} Representation are segmented points around each target. The attention of the person in red is shown on the static and dynamic scene.} }
\label{fig:model}
\end{figure*}
\subsection{Overview}
Our goal is a model that predicts accurate, diverse and safe, \ie collision-free, future trajectories for all relevant road users in a driving scene.
We denote the past trajectories of all $n$ road users participating in the scene with $\mathbf{X} = \{X_1, X_2,..., X_n\}$.
Each trajectory is a sequence of vectors $X_i = [\overline{x}_{i}^t]_{t=-T{obs}}^0$ for $T_{obs}$ past consecutive time steps.
Each element is a $\overline{x}_{i}^t = (\delta x_{i}^t, \delta y_{i}^t)$, a 2D coordinate displacement vector for agent $i$, with $\delta x_{i}^t = x_{i}^{t+1} - x_{i}^t$.
We also denote the ground truth future trajectories also containing 2D displacement vectors as elements with $\mathbf{Y} = \{Y_1, Y_2,..., Y_n\}$ and the predicted trajectories with $\mathbf{\hat{Y}}$. The time span of the prediction is $[T_0, \ldots, T_\text{pred}]$. To reduce notation clutter, we drop the subscript and superscripts $i$ and $t$ when they can be inferred from the context.

The proposed\model uses a conditional Generative Adversarial Network~\cite{gupta2018social}, as well as a Critic Network \cite{sutton2000policy}, \cite{sutton1984temporal},\cite{barto1984neuron}, to generate novel and safe future trajectories for the road users in the scene, see Figure~\ref{fig:model} for an overview.
Next, we describe the particular design of \model.

\subsection{Generating Trajectories with \model}
The conditional GAN of \model comprises of two modules, a \emph{Generator}, and a \emph{Discriminator}, as seen in the left of Figure ~\ref{fig:model}.
The generator of the conditional GAN is composed of two sub-networks.

The first sub-network is a $\theta$-parameterized encoder $\bfnot{u}=g^{enc}(\bfnot{X}; \theta)$ that receives as input the past trajectories of all road users and maps them to a common embedding space $\bfnot{u}$.
The purpose of the encoder is to aggregate the current state of all road users jointly. 
The aggregated state of all road users is then passed to the second sub-network, which is a $\omega$-parameterized decoder $\hat{\bfnot{Y}}=g^{dec}(\bfnot{u}, F_{d}, F_{s}, z;\omega)$, where $F_{d}, F_{s}$ are the dynamic and static aggregated features that will be detailed in \ref{section:attentive_scene}, and $z \sim N(0,1)$ is the random noise vector for sampling novel trajectories.
The purpose of the decoder is to generate novel future coordinates $\hat{\bfnot{Y}}$ for all $n$ road users simultaneously.

The generator is followed by the discriminator, which is implemented as a third $\phi$-parameterized sub-network $D = d(\bfnot{X}, \bfnot{Y}, \bfnot{\hat{Y}}| \phi)$.
The discriminator receives as inputs the past trajectories $\bfnot{X}$, as well as the ground truth future trajectories $\bfnot{Y}$ and the predicted trajectories $\bfnot{\hat{Y}}$.
Being adversaries, the generator and the discriminator are trained in an alternative manner \cite{goodfellow2014generative}: the generator generates gradually more  realistic trajectories, while the discriminator classifies which trajectories are fake thus motivating the generator to become better.

\subsection{\reward for Obtaining Safe Future Trajectories}

While the generator and the discriminator modules are trained to generate plausible future trajectories, the proposed trajectories are no provisions to avoid possible collisions in the predicted future trajectories.
To this end, we incorporate a \reward module responsible for minimizing the static and dynamic future collisions, as shown in the right of Figure~\ref{sec:method}.

A natural objective for the \reward module is a supervised loss that minimizes the number of future dynamic and static collisions.
Unfortunately, integrating such an objective to the model is not straightforward.
The number of non-colliding trajectories existing in data sets is overwhelmingly higher than for colliding trajectories.
Moreover, having a generative model means that the generated trajectories will not necessarily imitate the ground truth ones.

To this end, we rely on the off-policy actor-critic framework \cite{sutton1998introduction} to define the objective of \reward. 
The actor is the trajectory generator decoder $g^{dec}(\cdot)$, taking a displacement action $\hat{\bfnot{Y}}$.
The \reward then is a $\psi$-parameterized sub-network $v=o(\mathbf{\hat{Y}}, F_s, F_d; \psi)$ that predicts how likely an eminent collision is or not, given $\hat{\bfnot{Y}}$.
The action value $v$, is the score for the future action causing a collision.
Another choice would have been to simply return the reward signal to the generator.
However, in that case we would need to rely on high-variance gradients returned by REINFORCE.

\textbf{Generator} For training \model, we must take into account the opposing goals of the generator and the discriminator, as well as the regularizing effect of the critic.
Following standard GAN procedure we arrive at a min-max training objective for the generator and the discriminator \cite{goodfellow2014generative}. As the discriminator quickly learns to tell the real and generated samples apart, training the generator is harder than training the discriminator, leading to slow convergence or even failure \cite{salimans2016improved}.

Inspired from \cite{gupta2018social}, we introduce an additional auto-encoding loss to the ground truth for the best generated sample, $\mathbf{\hat{Y}}$, $\argmin_{\theta, \phi, \omega}  \sum_i \normeuc{\mathbf{Y}_i - \mathbf{\hat{Y}}_{i_{K}}}^2$. An additional benefit of the auto-encoding loss is that it encourages the model to avoid trivial solutions and mode collapse, as well as to increase the diversity of future generated trajectories.

\textbf{Critic} For the training of the \reward we employ a version of the Deep Deterministic Policy Gradient algorithm~\cite{lillicrap2015continuous}, more specifically, the DDPG variant~\cite{heess2015memory} that uses a sequence of observations. Regarding the training objective of the \reward sub-network, we simply opt for minimizing the squared error when a collision was predicted and when it occurred: \\
$
\argmin_{\psi} 
\mathbb{E}_{z\sim p_{data}(z)}\normeuc{o(\hat{\bfnot{Y}}, F_s, F_d; \psi) - R_\mathbf{\hat{Y}}},
$
where $R_\mathbf{\hat{Y}}$ indicates the presence or not of a collision and can be computed by any collision checking algorithm, here \cite{ericson2004real}.
It is important to note that following \cite{dasgupta_cvpr16, sadeghian2018sophie, gupta2018social}, all future predictions by the generator and the \reward are performed sequantially, one step at a time.
The Critic takes as input the generated trajectory from $g^{dec}$ up to time  $t$, while the collision checker algorithm computes $R_\mathbf{\hat{Y}}$ given the generated trajectory at the next time step, $t+1$.

%For the training of the \reward we employ a version of the Deep Deterministic Policy Gradient algorithm~\cite{lillicrap2015continuous}.
%Specifically, we rely on the DDPG variant~\cite{heess2015memory} that uses a sequence of observations and uses RNNs for both actor and critic networks.
The maximum reward is attained when the future trajectories cause minimum number of collisions.
In the end, the \reward sub-network acts as a regularizer to the rest of the network, guiding the generator to generate novel trajectories that cause mimimal collisions.

\subsection{Attentive Scene Representation}\label{section:attentive_scene}

To generate novel, accurate, diverse and safe future trajectories, a good representation of the scene as well as the rest of the agents in the scene is required. 
What is more, this representation must attend to the relevant elements in the vicinity of every agent.

To this end, we propose the following tensor representations $F_d, F_s$ for capturing possible dynamic interactions between the agents themselves, as well as the static interactions between the agents and the environment, respectively, see Fig. \ref{fig:model}.
Both representations for dynamic and static interactions follow the same design.
We center a grid around the $i$-th agent $G_i=\{g_{rc}\}$, where $r$ and $c$ run over the height and width of the grid.
We then check if a given grid location is occupied by either another agent or a static element.
If the $rc$ grid location is occupied by an agent $j$, we then set the respective entry in $F_{d, i}$ to the hidden state of that agent, namely $F_{d, i}^{rc}=h_j$.
Similarly for static features, if the $rc$ grid location is occupied by a static element we set the respective entry in $F_{s, i}$ to the one-hot vector representation of that static element.

Last, we employ an attention mechanism to prioritize certain elements in the latent state representations $F_d, F_s$ of the $i$-th agent.
The attention is applied spatially, such that specific coordinates on the grid receive higher importance than others.

\subsection{Architecture}

For the encoder we opt for a Long short-term memory (LSTM) network~\cite{gupta2018social}.
The observed trajectories are first embedded into a multilayer perceptron (MLP) in order to obtain a fixed length vector, $\bfnot{e} = \texttt{MLP}(\mathbf{X}; W_{emb})$, where $\mathbf{X}$ are the displacement coordinates of the pedestrians in the scene at all time steps $t=[-T_{obs}, ..., 0]$.
The LSTM then updates its hidden states sequentially.
The last hidden state of the LSTM is concatenated with the static and dynamic attentive representations $F_d, F_s$ and the noise input $z$ and then passed to the decoder.
For the decoder we rely also on a separate LSTM network that \emph{does not} share weights with the encoder.
The LSTM decoder generates the hidden states of each person for $t=[0,..., T_\text{pred}]$ predicted time steps.
The final hidden state is converted into a displacement vector and added onto the previous prediction $\hat{\bfnot{Y}}^{t-1}$, $\hat{\mathbf{Y}}^t = \hat{\mathbf{Y}}^{t-1} + \texttt{MLP}( \mathbf{h}_{dec}^{t} ; W_{T})$
where $W_{T}$ is a learnable transformation matrix.
The critic takes $K$ predictions $\hat{\bfnot{Y}}_i^{t}$ of agent $i$, together with the state in which that agent was $F_s^t, F_d^t$ and assigns a reward for each time-step. In this way, it learns to predict the likelihood of collision at $t+1$, which is dependant on the motion of agent $i$, but also of another agent $j$. If there are multiple approaching neighbors, chances of a collision are higher and since we want to reward avoiding collisions especially in these difficult situations, we do not normalize over the number of agents. We implement the Critic as an LSTM and an MLP and train it with Mean Squared Error between the number of collisions.

%
% \begin{equation}
% \begin{aligned}
% \hat{\mathbf{Y}}^t = \hat{\mathbf{Y}}^{t-1} + \texttt{MLP}( \mathbf{h}_{dec}^{t} ; W_{T})
% \end{aligned}
% \end{equation}
%

% %
% \begin{equation}
% \bfnot{c}_{enc}^{t}, \bfnot{h}_{enc}^{t} = \texttt{LSTM}(\bfnot{h}_{enc}^{t-1}, \bfnot{c}_{enc}^{t-1}, \bfnot{e}^t; W_{enc}).
% \end{equation}

% \begin{equation}
% \bfnot{h}_{dec}^{t}, \bfnot{c}_{dec}^{t} = \texttt{LSTM}(\bfnot{h}_{dec}^{t-1}, \bfnot{c}_{dec}^{t-1}, \bfnot{e}^t; W_{dec}).
% \end{equation}

\section{Experiments}\label{sec:results}
We evaluate our method on two state-of-the-art datasets used for future trajectory prediction of pedestrians.

\textbf{Data} UCY
\cite{leal2014learning} has 3 videos, recorded from a bird's eye point of view in 2 different outdoor locations.
It contains annotated positions for pedestrians only. The average length of a video is 361 seconds and it has 250 approximately people per video. Following\cite{dasgupta_cvpr16} we use the leave-one-out approach during training. 
The Stanford Drone Dataset \cite{robicquet2016learning} contains annotated positions of various types of agents that navigate inside a university campus. On average, there exist 104 agents per video.
We use the standard train and test splits provided by TrajNet \cite{sadeghiankosaraju2018trajnet}.

\textbf{Evaluation metrics} We use the standard metrics reported in the literature~\cite{dasgupta_cvpr16, gupta2018social, sadeghian2018sophie}, namely the minimum Average Displacement Error, $\texttt{mADE} = \min_{K} \frac{1}{T}\sum_{t=1}^{T} \normeuc{Y_{i}^t - \hat{Y}_{i}^t}$, and minimum Final Displacement Error, $\texttt{mFDE} = \min_{K} \normeuc{  Y_{i}^{T_{pred}} - \hat{Y}_{i}^{T_{pred}} }$ over best sample $K$. 

Last, to evaluate the capacity of the methods to avoid collisions, we measure the number of collisions with dynamic and static elements.
The number of collisions is computed as $\texttt{NC} = \sum_{i,j=1}^k  \sum_{t=1}^T\delta_{ij}\cdot\mathbf{1}(\hat{Y}_{i}^t-\hat{Y}_{j}^t < \epsilon)$
where $\delta_{ij}$ is the kronecker delta function, $\mathbf{1}$ is the indicator function, and $\epsilon = 0.10$ (here $i$ and $j$ are two different agents).

\textbf{Implementation details}
Following standard protocol ~\cite{sadeghiankosaraju2018trajnet} we observe trajectories for $T_\text{obs}=8$ times steps (3.2 seconds).
We then predict the next $T_\text{pred}=12$ (4.8 seconds) time steps.
The encoder and decoder LSTM have an embedding size of 32 each.
We  use batch normalization \cite{ioffe2015batch} and Adam optimization~ \cite{kingma2014adam} with learning rates $10^-3$ for the generator and $10^-4$ for the \reward and the discriminator.
Like~\cite{dasgupta_cvpr16} we bound and normalize distances between agents and the boundary points within a neighborhood.

\textbf{Baselines}.
We report comparisons with  state-of-the-art methods on the UCY and SDD: SocialGAN \cite{gupta2018social}, SocialLSTM \cite{dasgupta_cvpr16}, Car-Net \cite{sadeghian2017car}, SoPhie \cite{sadeghian2018sophie}, DESIRE \cite{lee2017desire}, as well as a linear regressor that estimates parameters by minimizing the least square error and Social Force implemented in \cite{sadeghian2018sophie}. 

\subsection{Evaluating attentive scene representation features}
First, we evaluate whether the proposed attentive scene representations allow for better generation of future trajectories.
In Table~\ref{table:1_2} we show results for trajectory generation when we turn on and off the proposed static and dynamic pooling, while also turning off the \reward module to minimize external influences. Following dataset conventions, errors are measured in meters for UCY and pixels for SDD. We observe that when switching on the proposed attentive scene representations we obtain lower errors in terms of \texttt{mADE} and \texttt{mFDE} for both UCY and SDD.
We conclude that the attentive scene representations allows for more accurate future trajectory generation.

\begin{table}[t]
\centering
\caption{
Minimum Average (ADE) and Final Displacement Error (FDE) for SafeCritic with and without Attentive Scene Representation (ASR). We conclude that the proposed interaction poolings lead to better modelling of the interactions for predicting future trajectorieslight increase of the speed. Table 2: Number of collisions between agents on UCY. Our model scores 0.01 vs 0.06 and 0.05 compared to LSTM, SocialLSTM  and SocialGAN on the UCYSDD.}
\begin{sc}
\begin{adjustbox}{max width=.48\linewidth}
\resizebox{\columnwidth}{!}{
\begin{tabular}{p{2cm}cc}
    \toprule
    \multirow{2}{5cm}{\textbf{Data}} 
    & \textbf{Without ASR} 
    & \textbf{With ASR} \\
            \cmidrule{2-3}    
                    & mADE / mFDE & mADE / mFDE \\
    \midrule       
    UCY             & 0.40 / 0.93 
                    & 0.39 / 0.84 \\
    % \midrule
    SDD             & 16.18 / 29.11 
                    & 15.06 / 28.45   \\
  \bottomrule
  \end{tabular}
  }
  \end{adjustbox}
\end{sc}
\label{tab:left}
\begin{sc}
\begin{adjustbox}{max width=.48\linewidth}
\resizebox{\columnwidth}{!}{
\begin{tabular}{p{2cm}ccccc}
    \toprule
    \textbf{Data}
                                    & \textbf{GT}
                                    & \textbf{Linear}
                                    & \textbf{SocialGAN}
                                    & \textbf{SoPhie}
                                    & \textbf{SafeCritic}
                                    \\
    \midrule
    % Students3                       & 0.12
    \textbf{UCY}                    \\
   Students3                        & 0.12
                                    & 1.24
                                    & 0.55
                                    & 0.62
                                    & 0.48
                                    \\ 
    % Zara1                           & \str{0.00??}
    Zara1                           & 0.00
                                    & 3.77
                                    & 1.74
                                    & 1.02
                                    & 0.18
                                    \\ 
    % Zara2                           & \underline{0.73}
    Zara2                           & 0.73
                                    & 3.63
                                    & 2.02
                                    & 1.46
                                    & 1.22
                                    \\
                                    \midrule
    % Average                         & \str{0.00 ??}
    Average                         & 0.28
                                    & 2.88
                                    & 1.43
                                    & \underline{1.03}
                                    & \textbf{0.62}
                                    \\ 
  \bottomrule
  \end{tabular}
  }
  \end{adjustbox}
\end{sc}
\label{table:1_2}
\end{table}
\addtocounter{table}{1}

\subsection{Evaluating the safety of \model}

Next, we examine how safe the generated trajectories are in terms of the number of collisions caused.
We report results in Table 2. A collision is defined with a threshold of 0.10m between two locations. The videos in SDD are less crowded and there are no people within a proximity of 0.10 m.
We make two observations.
Under all settings and data sets\model generates three times fewer collisions compared to SocialGAN, Sophie and a linear regressor.
What is more, \model is quite close to the number of the collisions that are actually observed in the ground truth trajectories.
This is important because collisions may actually happen in real life and thus a model should also generate collisions when these are likely to occur.

\subsection{Evaluating the accuracy of \model}

Next, we examine how accurate the generated future trajectories are.
We report results in Table~\ref{table:2a} for UCY and Table~\ref{table:2b} for SDD. We observe that \model generates more accurate future trajectories than all other baselines in UCY.
\begin{table*}[th!]
\caption{Given 8 past time steps the proposed SafeCritic is more accurate in predicting the future 12 timesteps on UCY (3 videos, 2492 frames).}
\centering
\begin{sc}
\begin{adjustbox}{max width=\textwidth}
% \resizebox{\columnwidth}{!}{
\begin{tabular}{p{2cm}cccccc}
    \toprule
    \multirow{2}{6cm}{\textbf{Data}} 
                                   & \textbf{SocialLSTM}
                                   & \textbf{SocialGAN}
                                    & \textbf{SoPhie}
                                   & \textbf{Linear}
                                   & \textbf{LSTM}
                                   & \textbf{SafeCritic}
                                   \\
                                   \cmidrule{2-7}
                                   & mADE / mFDE
                                   & mADE / mFDE
                                   & mADE / mFDE
                                   & mADE / mFDE
                                   & mADE / mFDE
                                   & mADE / mFDE
                                   \\
    \midrule       
% \midrule
    \textbf{UCY} \\
    Students3                      & 0.67 / 1.40
                                   & 0.76 / 1.52
                                   &  \underline{0.54} /  \underline{1.24}
                                   & 0.82 / 1.59
                                   & 0.61 / 1.31
                                   & \textbf{0.52} / \textbf{1.23}
                                   \\ 
    Zara1                          & 0.47 / 1.00
                                   & 0.35 / 0.68
                                   &  \underline{0.30} /  \underline{0.63}
                                   & 0.62 / 1.21
                                   & 0.41 / 0.88
                                   & \textbf{0.29} / \textbf{0.60}
                                   \\ 
    Zara2                          & 0.56 / 1.17
                                   & 0.42 / 1.01
                                   &  \underline{0.38} /  \underline{0.78}
                                   & 0.77 / 1.48
                                   & 0.52 / 1.11
                                   & \textbf{0.30} / \textbf{0.69}
                                   \\
                                   \midrule
    Average                        & 0.56 / 1.19
                                   & 0.51 / 0.96
                                   &  \underline{0.41} /   \underline{0.88}
                                   & 0.74 / 1.43
                                   & 0.51 / 1.10
                                   & \textbf{0.37} / \textbf{0.84}
                                   \\
                                   \midrule
  \end{tabular}
%   }
  \end{adjustbox}
\end{sc}
\label{table:2a}
\end{table*}

Importantly, \model comfortably outperforms all methods on the much larger and complex SDD dataset.
We also evaluate the diversity of the generated trajectories in Fig. ~\ref{fig:diverse}.
We conclude that \model predicts future trajectories that are not only accurate, but also cause much fewer collisions and that are more diverse.
\begin{table*}[th!]
\caption{Given 8 past time steps the proposed SafeCritic is more accurate in predicting the future 12 timesteps on SDD (48 videos, 14558 frames).}
\centering
\begin{sc}
\begin{adjustbox}{max width=\textwidth}
% \resizebox{\columnwidth}{!}{
\begin{tabular}{p{2cm}cccccccc}
    \toprule
    \multirow{2}{6cm}{\textbf{Data}} 
                                   & \textbf{SocialLSTM}
                                   & \textbf{SocialGAN}
                                    & \textbf{SoPhie}
                                   & \textbf{Desire}
                                   & \textbf{Linear}
                                   & \textbf{SF}
                                   & \textbf{CAR-Net}
                                   & \textbf{SafeCritic}
                                   \\
                                   \cmidrule{2-9}
                                   & mADE / mFDE
                                   & mADE / mFDE
                                   & mADE / mFDE
                                   & mADE / mFDE
                                   & mADE / mFDE
                                   & mADE / mFDE
                                   & mADE / mFDE
                                   & mADE / mFDE
                                   \\
%    \midrule       
% \midrule

                                   \midrule
    \textbf{SDD}             & 31.19 / 56.97
                                  & 27.25 / 41.44
                                  &  \underline{16.27} /  \underline{29.38}
                                  & 19.25 / 34.05
                                  & 37.11 / 63.51
                                  & 36.48 / 58.14
                                  & 25.72 / 51.80
                                  & \textbf{14.39} / \textbf{27.39}
                                  \\
  \bottomrule
  \end{tabular}
%   }
  \end{adjustbox}
\end{sc}
\label{table:2b}
\end{table*}

\subsection{Qualitative results}
\begin{figure}[th!]
%\begin{subfigure}[a]{.49\textwidth}
\centering
    \begin{minipage}[t]{.28\textwidth}
        \centering
        \includegraphics[width=\textwidth,trim={5 5 5 5},clip]{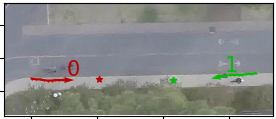}
        %Observed
    \end{minipage}  
    \begin{minipage}[t]{.28\textwidth}
        \centering
        \includegraphics[width=\textwidth,trim={5 5 5 5},clip]{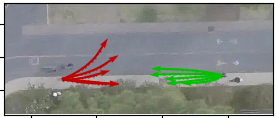}
        %SafeCritic
    \end{minipage}  
    \begin{minipage}[t]{.20\textwidth}
        \centering
        \includegraphics[width=\textwidth,trim={5 75 5 29},clip]{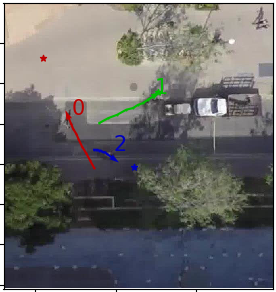}
        %Observed
    \end{minipage}
    \begin{minipage}[t]{.20\textwidth}
        \centering
        \includegraphics[width=\textwidth,trim={5 70 5 26},clip]{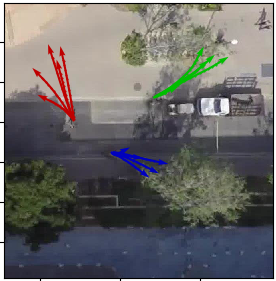}
        %SafeCritic
    \end{minipage}  
    % \caption{HYANG}
  \label{fig:zara1}
%\end{subfigure}
\caption{As expected SafeCritic generates diverse trajectories. Five trajectories of two persons walking towards a goal (star) in the observed (left) and predicted (right) trajectories.}
% \vspace{-1em}
\label{fig:diverse}
\end{figure}

Last, we present some qualitative results for \model. In Fig.~\ref{fig:diverse} we plot various generated trajectories at a specific moment in time, examining whether mode collapse is happening.
We observe that \model does not suffer from mode collapse~\cite{salimans2016improved}, as also quantified earlier.

\begin{figure}[th!]

\centering
%    \begin{minipage}[t]{.26\textwidth}
%        \centering
%        \includegraphics[width=\textwidth,trim={0 2cm 1cm 1cm},clip]{images/attention/frame_243_obs_sample_0.jpg}
%        Observed
%    \end{minipage}
    \begin{minipage}[t]{.24\textwidth}
        \centering
        \includegraphics[width=\textwidth,trim={0 2.3cm 1cm 1.7cm},clip]{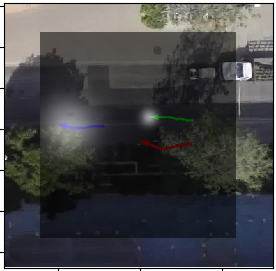}
        %Attention
    \end{minipage}  
        \begin{minipage}[t]{.24\textwidth}
        \centering
        \includegraphics[width=\textwidth,trim={0 2.6cm 1cm 1.7cm},clip]{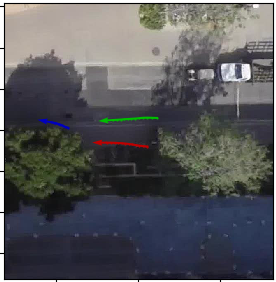}
        %Prediction
    \end{minipage} 
    \begin{minipage}[t]{.24\textwidth}
        \centering
        \includegraphics[width=\textwidth,trim={2.5cm 6.2cm 0 0.2cm},clip]{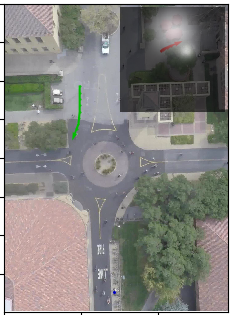}
        %Attention
    \end{minipage}  
        \begin{minipage}[t]{.24\textwidth}
        \centering
        \includegraphics[width=\textwidth,trim={2.5cm 6.2cm 0 0.2cm},clip]{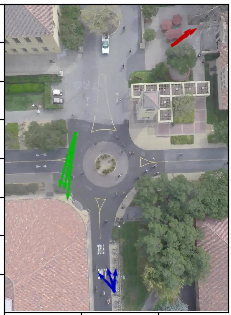}
        %Prediction
    \end{minipage} 
    % \caption{HYANG}
  \label{fig:zara1dd}
%\end{subfigure}
\caption{The attention (left) of a person (red) show the importance of other agents (blue and green) (left columns) and of nearby trees and buildings (right columns), resulting in alternating his trajectory (right).} 
% \vspace{-1em}
\label{fig:attention_static}
\end{figure}

\begin{figure}[th!]
%\begin{subfigure}[a]{.49\textwidth}
\centering
    \begin{minipage}[t]{.16\textwidth}
        \centering
        \includegraphics[width=\textwidth,trim={4cm 7cm 19cm 5cm},clip]{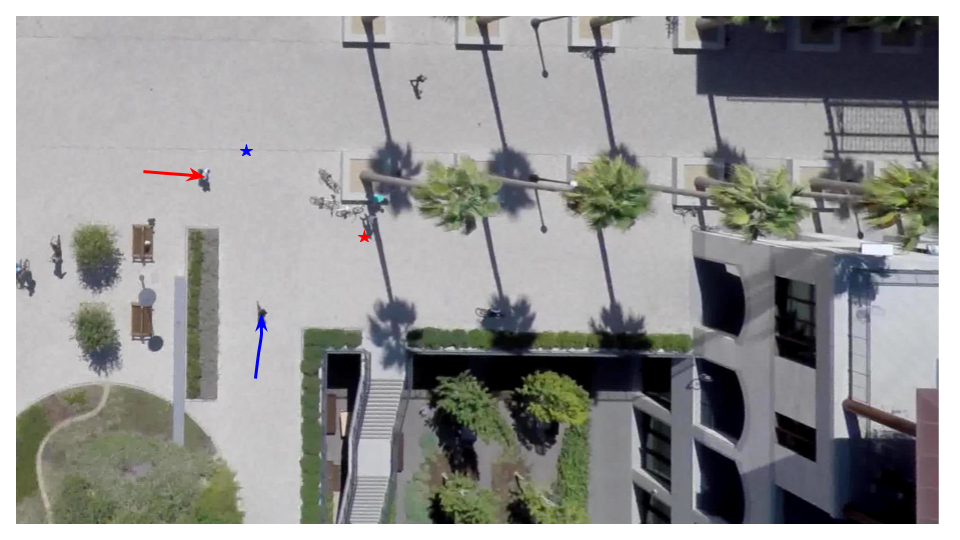}
     \small{Observed}
    \end{minipage}
    \begin{minipage}[t]{.16\textwidth}
        \centering
        \includegraphics[width=\textwidth,trim={4cm 7cm 19cm 5cm},clip]{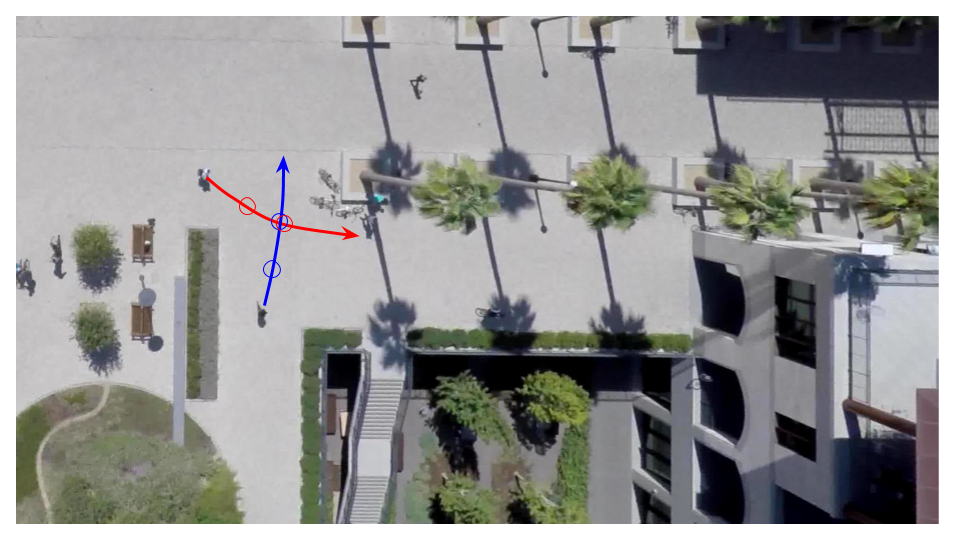}
         \small{SocialGAN}
    \end{minipage} 
        \begin{minipage}[t]{.16\textwidth}
        \centering
        \includegraphics[width=\textwidth,trim={4cm 7cm 19cm 5cm},clip]{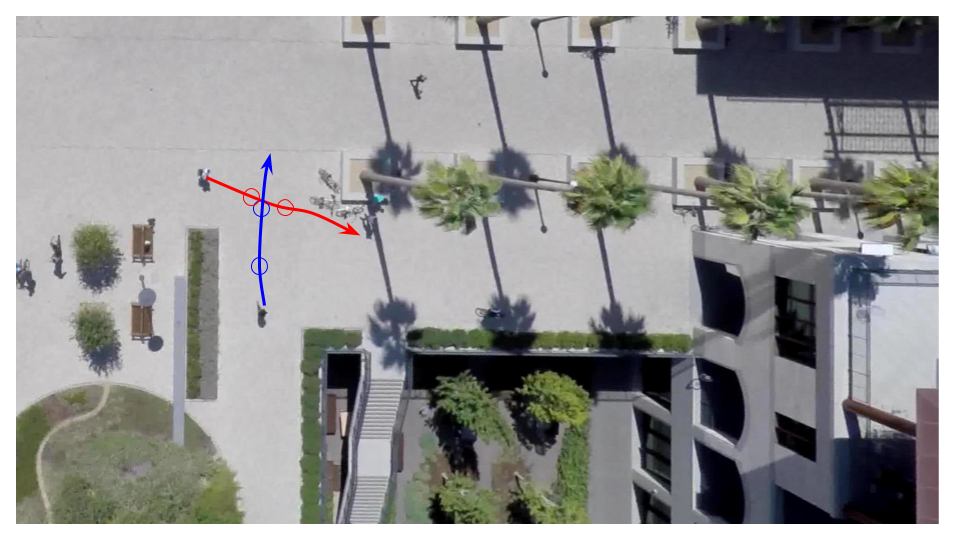}
        \small{SafeCritic}
    \end{minipage} 
    \begin{minipage}[t]{.16\textwidth}
        \centering
        \includegraphics[width=\textwidth,trim={6cm 8cm 17cm 4cm},clip]{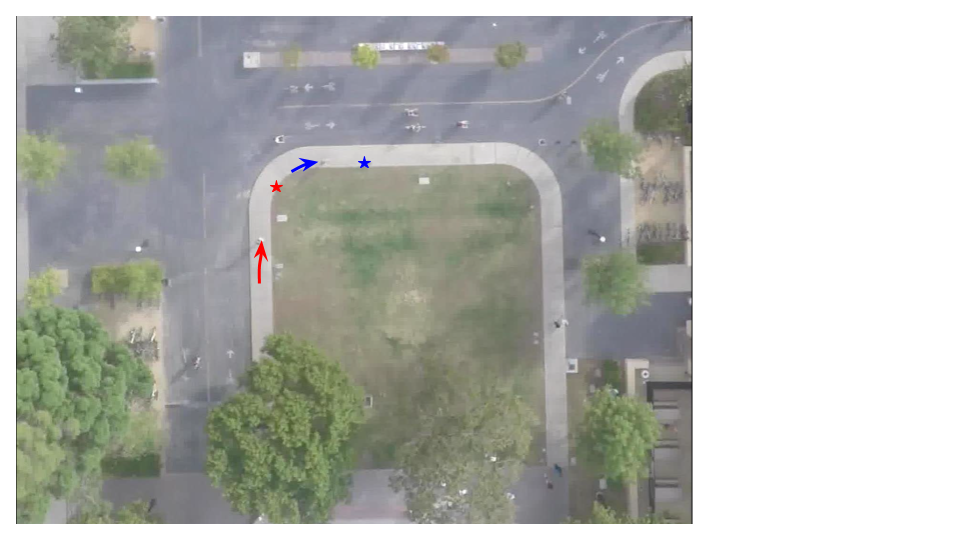}
        \small{Observed}
    \end{minipage}
    \begin{minipage}[t]{.16\textwidth}
        \centering
        \includegraphics[width=\textwidth,trim={6cm 8cm 17cm 4cm},clip]{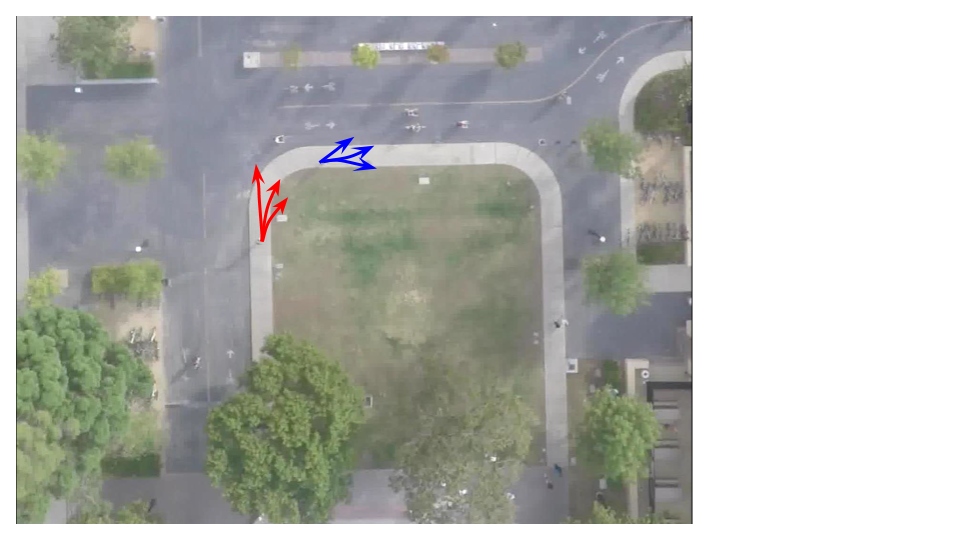}
        \small{SocialGAN}
    \end{minipage}  
    \begin{minipage}[t]{.16\textwidth}
        \centering
        \includegraphics[width=\textwidth,trim={6cm 8cm 17cm 4cm},clip]{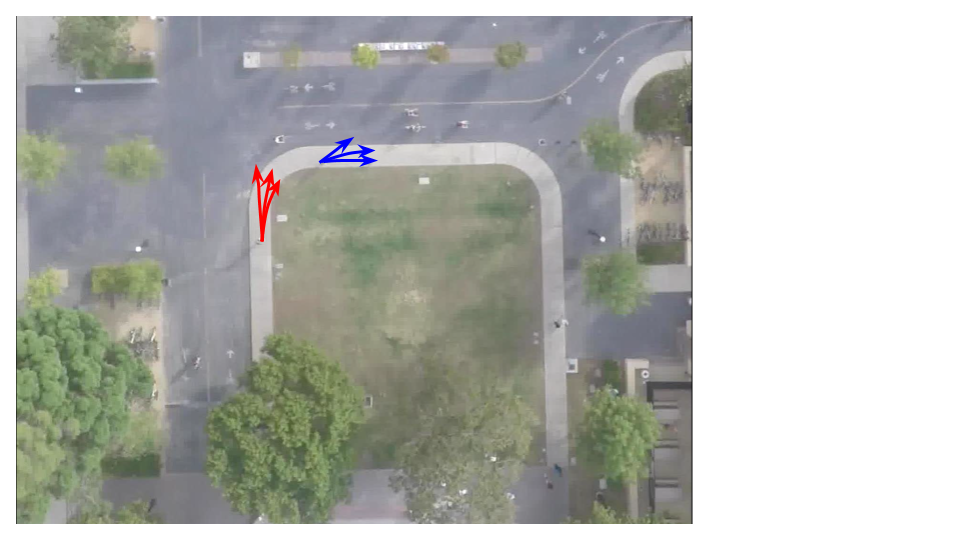}
        \small{SafeCritic}
    \end{minipage} 
    % \caption{HYANG}
  \label{fig:zara1l}
%\end{subfigure}
\vspace{10pt} 
\caption{As expected SafeCritic generates safe trajectories. Trajectories of two walking persons with final ground truth location (star). Collisions between people for Coupa 0 (left columns) with the locations at t=1s and 2s (circles) and going onto lawn for Bookstore 2 (right columns). }
% \vspace{-1em}
\label{fig:safe}
\end{figure}

Further, in Fig.~\ref{fig:attention_static} we show that the generated future trajectories are nicely sampled such that the attended agents and static elements are avoided and no collisions are generated. 
Last, in Fig. ~\ref{fig:safe} we show how the predicted locations of SafeCritic do not collide. The second row shows how multiple sampled trajectories of SafeCritic move around a lawn.

%Last, we present some qualitative results for \model. In Fig.~\ref{fig:diverse} we plot various generated trajectories at a specific moment in time, examining whether mode collapse is happening. Mode collapse is problematic, because the model could learn to generate trajectories that cause a low number of collisions, while failing to predict the whole space of possible future paths. We observe that \model does not suffer from mode collapse~\cite{salimans2016improved}, as also quantified earlier. Interestingly, the diversity of the generated trajectories is different for different pedestrians, showcasing the important of the attentive scene representations. Further, in Fig.~\ref{fig:attention_static} we visualize the attention to other agents and static scene elements, like nearby trees and buildings. We observe that the generated future trajectories are nicely sampled such that the attended agents and static elements are avoided and no collisions are generated. Last, in Fig. ~\ref{fig:safe} we show how the predicted locations of SafeCritic do not collide. The second row shows how multiple sampled trajectories of SafeCritic move around a lawn.

\section{Conclusion}
In this work we introduced\model, a model that is based on training in an adversarial way along with a reinforcement learning based reward network to predict both accurate and safe trajectories. A novel way of handling static and dynamic features in both the generator and reward network makes it possible to discard unsafe trajectories. Our proposed collision metric enables the evaluation of safety aspect of predicted trajectories. We show that \model outperforms baseline models both on accuracy as well as on the number of collisions. Furthermore, our experiments show that a convex combination of trajectory's `safe' and `real' aspects achieves a perfect balance for the final prediction.

{\small
\bibliographystyle{unsrt}
\bibliography{main}

\begin{thebibliography}{10}

\bibitem{gupta2018social}
Agrim Gupta, Justin Johnson, Li~Fei-Fei, Silvio Savarese, and Alexandre Alahi.
\newblock Social gan: Socially acceptable trajectories with generative
  adversarial networks.
\newblock In {\em IEEE Conference on Computer Vision and Pattern Recognition
  (CVPR)}, number CONF, 2018.

\bibitem{sadeghian2018sophie}
Amir Sadeghian, Vineet Kosaraju, Ali Sadeghian, Noriaki Hirose, and Silvio
  Savarese.
\newblock Sophie: An attentive gan for predicting paths compliant to social and
  physical constraints.
\newblock {\em arXiv preprint arXiv:1806.01482}, 2018.

\bibitem{lee2017desire}
Namhoon Lee, Wongun Choi, Paul Vernaza, Christopher~B Choy, Philip~HS Torr, and
  Manmohan Chandraker.
\newblock Desire: Distant future prediction in dynamic scenes with interacting
  agents.
\newblock In {\em Proceedings of the IEEE Conference on Computer Vision and
  Pattern Recognition}, pages 336--345, 2017.

\bibitem{goodfellow2014generative}
Ian Goodfellow, Jean Pouget-Abadie, Mehdi Mirza, Bing Xu, David Warde-Farley,
  Sherjil Ozair, Aaron Courville, and Yoshua Bengio.
\newblock Generative adversarial nets.
\newblock In {\em Advances in neural information processing systems}, pages
  2672--2680, 2014.

\bibitem{IRL}
Andrew~Y. Ng and Stuart~J. Russell.
\newblock Algorithms for inverse reinforcement learning.
\newblock In {\em ICML}, 2000.

\bibitem{dasgupta_cvpr16}
Alexandre Alahi, Kratarth Goel, Vignesh Ramanathan, Alexandre Robicquet,
  Li~Fei-Fei, and Silvio Savarese.
\newblock Social lstm: Human trajectory prediction in crowded spaces.
\newblock In {\em Proceedings of the IEEE International Conference on Computer
  Vision and Pattern Recognition}, 2016.

\bibitem{gauthier2014conditional}
Jon Gauthier.
\newblock Conditional generative adversarial nets for convolutional face
  generation.
\newblock {\em Class Project for Stanford CS231N: Convolutional Neural Networks
  for Visual Recognition, Winter semester}, 2014(5):2, 2014.

\bibitem{mirza2014conditional}
Mehdi Mirza and Simon Osindero.
\newblock Conditional generative adversarial nets.
\newblock {\em arXiv preprint arXiv:1411.1784}, 2014.

\bibitem{rhinehart2018r2p2}
Nicholas Rhinehart, Kris~M Kitani, and Paul Vernaza.
\newblock R2p2: A reparameterized pushforward policy for diverse, precise
  generative path forecasting.
\newblock In {\em Proceedings of the European Conference on Computer Vision
  (ECCV)}, pages 772--788, 2018.

\bibitem{vemula2018social}
Anirudh Vemula, Katharina Muelling, and Jean Oh.
\newblock Social attention: Modeling attention in human crowds.
\newblock In {\em 2018 IEEE International Conference on Robotics and Automation
  (ICRA)}, pages 1--7. IEEE, 2018.

\bibitem{helbing1995social}
Dirk Helbing and Peter Molnar.
\newblock Social force model for pedestrian dynamics.
\newblock {\em Physical review E}, 51(5):4282, 1995.

\bibitem{sud2008real}
Avneesh Sud, Russell Gayle, Erik Andersen, Stephen Guy, Ming Lin, and Dinesh
  Manocha.
\newblock Real-time navigation of independent agents using adaptive roadmaps.
\newblock In {\em ACM SIGGRAPH 2008 classes}, page~56. ACM, 2008.

\bibitem{trautman2010unfreezing}
Peter Trautman and Andreas Krause.
\newblock Unfreezing the robot: Navigation in dense, interacting crowds.
\newblock In {\em Intelligent Robots and Systems (IROS), 2010 IEEE/RSJ
  International Conference on}, pages 797--803. IEEE, 2010.

\bibitem{becker2018evaluation}
Stefan Becker, Ronny Hug, Wolfgang H{\"u}bner, and Michael Arens.
\newblock An evaluation of trajectory prediction approaches and notes on the
  trajnet benchmark.
\newblock {\em arXiv preprint arXiv:1805.07663}, 2018.

\bibitem{sadeghiankosaraju2018trajnet}
Amir Sadeghian, Vineet Kosaraju, Agrim Gupta, Silvio Savarese, and Alexandre
  Alahi.
\newblock Trajnet: Towards a benchmark for human trajectory prediction.
\newblock {\em arXiv preprint}, 2018.

\bibitem{finn2016gan_irl}
Chelsea Finn, Paul~F. Christiano, Pieter Abbeel, and Sergey Levine.
\newblock A connection between generative adversarial networks, inverse
  reinforcement learning, and energy-based models.
\newblock {\em CoRR}, abs/1611.03852, 2016.

\bibitem{de2018molgan}
Nicola De~Cao and Thomas Kipf.
\newblock Molgan: An implicit generative model for small molecular graphs.
\newblock {\em arXiv preprint arXiv:1805.11973}, 2018.

\bibitem{guimaraes2017objective}
Gabriel~Lima Guimaraes, Benjamin Sanchez-Lengeling, Carlos Outeiral, Pedro
  Luis~Cunha Farias, and Al{\'a}n Aspuru-Guzik.
\newblock Objective-reinforced generative adversarial networks (organ) for
  sequence generation models.
\newblock {\em arXiv preprint arXiv:1705.10843}, 2017.

\bibitem{konda2000actor}
Vijay~R Konda and John~N Tsitsiklis.
\newblock Actor-critic algorithms.
\newblock In {\em Advances in neural information processing systems}, pages
  1008--1014, 2000.

\bibitem{sutton2000policy}
Richard~S Sutton, David~A McAllester, Satinder~P Singh, and Yishay Mansour.
\newblock Policy gradient methods for reinforcement learning with function
  approximation.
\newblock In {\em Advances in neural information processing systems}, pages
  1057--1063, 2000.

\bibitem{sutton1984temporal}
Richard~Stuart Sutton.
\newblock Temporal credit assignment in reinforcement learning.
\newblock 1984.

\bibitem{barto1984neuron}
AG~Barto.
\newblock Neuron-like adaptive elements that can solve difficult learning
  control-problems.
\newblock {\em Behavioural Processes}, 9(1), 1984.

\bibitem{sutton1998introduction}
Richard~S Sutton and Andrew~G Barto.
\newblock {\em Introduction to reinforcement learning}, volume 135.
\newblock MIT press Cambridge, 1998.

\bibitem{salimans2016improved}
Tim Salimans, Ian Goodfellow, Wojciech Zaremba, Vicki Cheung, Alec Radford, and
  Xi~Chen.
\newblock Improved techniques for training gans.
\newblock In {\em Advances in Neural Information Processing Systems}, pages
  2234--2242, 2016.

\bibitem{lillicrap2015continuous}
Timothy~P Lillicrap, Jonathan~J Hunt, Alexander Pritzel, Nicolas Heess, Tom
  Erez, Yuval Tassa, David Silver, and Daan Wierstra.
\newblock Continuous control with deep reinforcement learning.
\newblock {\em arXiv preprint arXiv:1509.02971}, 2015.

\bibitem{heess2015memory}
Nicolas Heess, Jonathan~J Hunt, Timothy~P Lillicrap, and David Silver.
\newblock Memory-based control with recurrent neural networks.
\newblock {\em arXiv preprint arXiv:1512.04455}, 2015.

\bibitem{ericson2004real}
Christer Ericson.
\newblock {\em Real-time collision detection}.
\newblock CRC Press, 2004.

\bibitem{leal2014learning}
Laura Leal-Taix{\'e}, Michele Fenzi, Alina Kuznetsova, Bodo Rosenhahn, and
  Silvio Savarese.
\newblock Learning an image-based motion context for multiple people tracking.
\newblock In {\em Proceedings of the IEEE Conference on Computer Vision and
  Pattern Recognition}, pages 3542--3549, 2014.

\bibitem{robicquet2016learning}
Alexandre Robicquet, Amir Sadeghian, Alexandre Alahi, and Silvio Savarese.
\newblock Learning social etiquette: Human trajectory understanding in crowded
  scenes.
\newblock In {\em European conference on computer vision}, pages 549--565.
  Springer, 2016.

\bibitem{ioffe2015batch}
Sergey Ioffe and Christian Szegedy.
\newblock Batch normalization: Accelerating deep network training by reducing
  internal covariate shift.
\newblock {\em arXiv preprint arXiv:1502.03167}, 2015.

\bibitem{kingma2014adam}
Diederik~P Kingma and Jimmy Ba.
\newblock Adam: A method for stochastic optimization.
\newblock {\em arXiv preprint arXiv:1412.6980}, 2014.

\bibitem{sadeghian2017car}
Amir Sadeghian, Ferdinand Legros, Maxime Voisin, Ricky Vesel, Alexandre Alahi,
  and Silvio Savarese.
\newblock Car-net: Clairvoyant attentive recurrent network.
\newblock {\em arXiv preprint arXiv:1711.10061}, 2017.

\end{thebibliography}
}

\end{document}